\documentclass[twoside]{article}

\usepackage[accepted]{style_files/aistats2020}
\usepackage{style_files/seb}

\begin{document}

%
\runningtitle{Radial Bayesian Neural Networks}

%

\twocolumn[

\aistatstitle{Radial Bayesian Neural Networks: Beyond Discrete Support In Large-Scale Bayesian Deep Learning}
\aistatsauthor{ Sebastian Farquhar* \And Michael A. Osborne \And Yarin Gal}
\aistatsaddress{OATML, University of Oxford \And MLRG, University of Oxford \And OATML, University of Oxford}
 ]

\begin{abstract}
  We propose Radial Bayesian Neural Networks (BNNs): a variational approximate posterior for BNNs which scales well to large models.
  Unlike scalable Bayesian deep learning methods like deep ensembles that have discrete support (assign exactly zero probability almost everywhere in weight-space) Radial BNNs maintain full support: letting them act as a prior for continual learning and avoiding the \textit{a priori} implausibility of discrete support.
  Our method avoids a sampling problem in mean-field variational inference (MFVI) caused by the so-called `soap-bubble' pathology of multivariate Gaussians.
  We show that, unlike MFVI, Radial BNNs are robust to hyperparameters and can be efficiently applied to challenging real-world tasks without needing ad-hoc tweaks and intensive tuning: on a real-world medical imaging task Radial BNNs outperform MC dropout and deep ensembles.

\end{abstract}

\section{INTRODUCTION}
\begin{figure}[t]
\vspace{-2mm}
\centering
\includegraphics{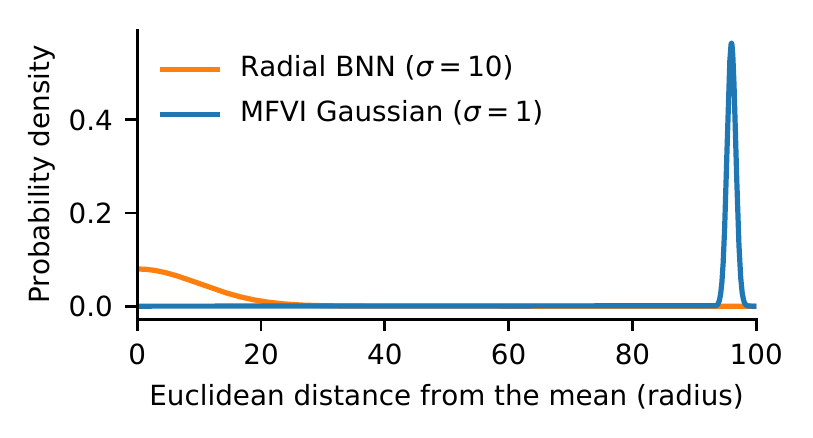}
\vspace{-7mm}
\caption{MFVI uses a multivariate Gaussian approximate posterior whose probability mass is tightly clustered at a fixed radius from the mean depending on the number of dimensions---the `soap-bubble'. In our Radial BNN, samples from the approximate posterior are more reflective of the mean. This helps training by reducing gradient variance. (Plotted p.d.f. is based on dimensionality of a 3x3 conv layer with 64 channels.)}
    \label{fig:pdf_plot}
    \vspace{-4mm}
\end{figure}

The most effective scalable methods for Bayesian deep learning have a significant shortcoming: they learn an approximate posterior distribution that has discrete support over the weight-space---the probability assigned to almost all possible weights is exactly zero.
This is true of methods like MC dropout \citep{gal_dropout_2015}, but also to samples from stochastic gradient Markov Chain Monte Carlo (MCMC) \citep{welling_bayesian_2011}, or deep ensembles \citep{lakshminarayanan_simple_2016} where finitely many samples appear in the empirical distribution.
This is implausible \textit{a priori}: from an epistemic perspective assigning zero posterior probability almost everywhere is pathological overconfidence.
But overconfidence is also unhelpful, these distributions are unsuitable as a data-dependent prior in continual learning---once a prior is exactly zero, no amount of data can update it.

Some variational inference methods \textit{do} learn approximate posteriors with full support over the weight-space.
`Mean-field' variational inference (which assumes independent weight distributions) is fast and has linear time complexity in the number of parameters \citep{hinton_keeping_1993, graves_practical_2011, blundell_weight_2015}.
Unfortunately, MFVI struggles in practice for tasks larger than roughly the scale of MNIST and is sensitive to hyperparameters \citep{wu_deterministic_2019}.
Tuning hyperparameters is a barrier to using MFVI for larger models where each iteration could take days.
To make MFVI work, researchers often resort to ad-hoc tweaks to the loss or optimization process which side-step the variational inference arguments that motivate the approach in the first place! (See \S\ref{s:tweaks}.)
Other research relaxes MFVI's independence assumption by introducing expensive techniques which are tractable for small networks with only thousands of parameters and low-dimensional problems (e.g., MNIST) \citep{louizos_structured_2016, sun_learning_2017, sun_functional_2019, oh_radial_2019}.
\textbf{What is missing is robust, scalable inference for BNNs that maintains full support.}

In this paper we identify a sampling problem at the heart of MFVI's failures---typical samples from the multivariate Gaussian approximate posterior used in MFVI are unrepresentative of the most-probable weights, and this problem gets worse for larger networks \citep{bishop_introduction_2006}.
Probability mass in a multivariate Gaussian is clustered in a narrow `soap-bubble' far from the mean (see Figure \ref{fig:pdf_plot}).\footnote{We refer readers to the Appendix \ref{a:soap_bubble} for more detail on this phenomenon. Intuitively, the issue arises because the space expands with the polynomial $r^{D}$ in the radius $r$ in $D$ dimensions, while the p.d.f. of the Gaussian falls exponentially. At the origin, the polynomial term is small, at infinity the exponential term is small, and almost all the probability mass lies in a narrow band in between.}
Unless the approximate posterior distribution is very tight, samples tend to be distant from each other.
This leads to exploding gradient variance whenever the posterior becomes broad, and prevents MFVI from actually fitting to the loss.
We demonstrate this in \S\ref{s:understanding}.

Therefore, we propose an alternative approximate posterior distribution without a `soap-bubble'.
The Radial BNN defines a simple approximate posterior distribution in a hyperspherical space corresponding to each layer, and then transforms this distribution into the coordinate system of the weights.
The typical samples from this distribution tend to come from areas of high probability density.
We show that the Radial BNN can be sampled efficiently in weight-space, without needing explicit coordinate transformations, and derive an analytic expression for the loss that makes training as fast and as easy to implement as MFVI.

\begin{figure}
    \centering
    \minipage{0.4\linewidth}
      \includegraphics[width=\linewidth]{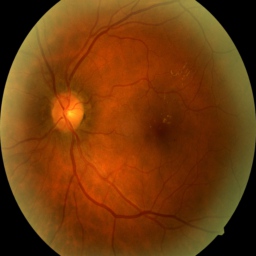}
    \endminipage \
    \minipage{0.4\linewidth}
      \includegraphics[width=\linewidth]{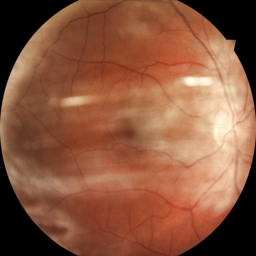}
    \endminipage \hfill
    
    \minipage{0.4\linewidth}
      \includegraphics[width=\linewidth]{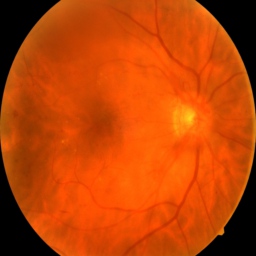}
    \endminipage \
    \minipage{0.4\linewidth}%
      \includegraphics[width=\linewidth]{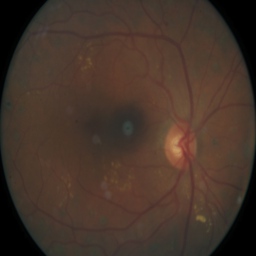}
    \endminipage
    \caption{Examples from retinopathy dataset. Top L: healthy eye. Top R: healthy eye with camera artefacts. Bottom: diseased eyes. The chance that bad images cause misdiagnosis makes uncertainty-aware models vital. Input dimension 334x bigger than MNIST.}
    \label{fig:eye_examples}
    \end{figure}

We establish the robustness and performance of Radial BNNs using a Bayesian medical imaging task identifying diabetic retinopathy in `fundus' eye images \citep{leibig_leveraging_2017}, using models with $\sim$15M parameters and inputs with $\sim$230,000 dimensions, in \S\ref{s:retinopathy} (see Figure \ref{fig:eye_examples}).
Radial BNNs are more robust to hyperparameter choice than MFVI and that Radial BNNs outperform the current state-of-the-art Monte-Carlo (MC) dropout and deep ensembles on this task.

In addition, because Radial BNN approximate posteriors have \textit{full support} over the weight-space, they can be used as a prior for further inference.
We show this in \S\ref{s:vcl} using a continual learning setting \citep{kirkpatrick_overcoming_2017, nguyen_variational_2018}, where a sequence of approximate posteriors are used as a prior to avoid catastrophic forgetting.
While we do not solve continual learning, we use the problem setting to demonstrate the potential to find rich data-dependent priors.
\section{PRIOR WORK}
Instead of point estimates, Bayesian neural networks (BNNs) place a parameterized distribution over each weight in a neural network \citep{mackay_practical_1992, neal_bayesian_1995}.
Many efficient approximations have been proposed to estimate the posterior distribution over those weights including mean-field variational inference \citep{hinton_keeping_1993, graves_practical_2011, blundell_weight_2015}, Monte Carlo (MC) dropout \citep{gal_dropout_2015}, stochastic gradient Markov Chain MC \citep{welling_bayesian_2011} and others.
Deep ensembles \citep{lakshminarayanan_simple_2016} have also been proposed as a way to learn a distribution over the weights (although the connection to the posterior remains unclear).

Unfortunately, existing robust methods, scalable to large models and datasets, do not learn a posterior with full support over the weight-space.
Monte Carlo dropout learns a parameterized distribution based on Bernouilli random variables, which only represent discrete points in weight-space.
Methods like deep ensembles and SG-MCMC instead produce a finite number of samples and estimate the predictive distribution using the an empirical distribution with discrete support.
These sorts of discrete distributions are epistemically pathological---they represent implausible overconfidence.
This makes them unsuitable as a prior for further inference.

Mean-field variational inference offers a fully supported distribution over the weights.
It sets an approximate posterior distribution, $q_{\thet}$, over each weight, $\mathbf{w}$, in the network---an independent Gaussian (where $\thet$ is $\{\boldsymbol{\mu} \cup \boldsymbol{\sigma}\}$).
It then optimizes a lower-bound on the marginal likelihood which tries to find the approximate posterior with the smallest KL-divergence to the true posterior.
This loss can be interpreted as balancing predictive accuracy on the data ($\mathbf{y}$ and $\mathbf{X}$), the entropy of the posterior, and the cross-entropy between the prior and posterior.
In the common case of a unit multivariate Gaussian prior and an approximate posterior $\mathcal{N}(\mu_i, \sigma_i^2)$ over the weights $w_i$, the negative evidence lower bound (ELBO) objective is:
\begin{align}
    \mathcal{L}_{\text{MFVI}} = &\overbrace{\rule{0pt}{3ex}\sum_i \frac{1}{2}\big[\sigma^2_i + \mu^2_i\big]}^{\text{prior cross-entropy}} - \overbrace{\rule{0pt}{3ex}\sum_i \log[\sigma_i]}^{\mathclap{\substack{\text{approximate-} \\ \text{posterior entropy}}}} \nonumber\\ &- \overbrace{\rule{0pt}{3ex}\E_{\mathbf{w} \sim q_{\theta}(\mathbf{w})} \big[ \log p(\mathbf{y}|\mathbf{w},\mathbf{X})\big]}^{\text{data likelihood}}. \label{eq:MFVI_unit_prior}
\end{align}

In practice, training BNNs with MFVI is difficult.
For example, \citet{wu_deterministic_2019} argue that it is sensitive to initialization and priors.
Others worry that the mean-field approximation is too constraining.
\citet{louizos_structured_2016, sun_learning_2017} and \citet{oh_radial_2019} have all introduced richer variational distributions which permit correlations between weights to be learned by the BNN; \citet{sun_functional_2019} instead perform inference in function-space.
Unfortunately, these methods are considerably more computationally expensive than MFVI and have only been demonstrated on problems \textit{at MNIST scale or below}.
\citet{wu_deterministic_2019} instead see the variance of ELBO estimates as the problem and introduce a deterministic alternative.
We agree that this is a crucial problem, but offer a simpler and cheaper alternative solution which does not require extra assumptions about the distribution of activations.
Note that \cite{osawa_practical_2019} present scalable inference for MFVI using variational online Gauss-Newton methods \citep{khan_fast_2018}.
However, this method relies on several significant approximations to make estimates of the Hessian tractable and the performance of the method lags significantly behind deep ensembles, which we compare to.

We note that there is a superficial similarity between our method and \citet{oh_radial_2019}, insofar as they also make use of a hyperspherical coordinate system for variational inference.
However, they use this coordinate system over each row in their weight matrix, rather than the whole layer, and introduce an expensive posterior distribution (von Mises-Fisher) to explicitly model weight correlations within rows, whereas we do \textit{not} seek to learn any correlations between parameters in the hyperspherical space.
That is, their method uses a different technique to solve a different problem.

\section{METHOD}\label{s:method}
A well-known property of multivariate Gaussians in high dimensions is that the probability mass concentrates in a `soap-bubble'---a narrow shell at a radius determined by the variance and the number of dimensions (e.g., \citep{bishop_introduction_2006, betancourt_conceptual_2018}).
This has the consequence that almost all samples from the distribution are very distant from the mean.
All else equal, we might expect this to lead to predictions and losses from multiple samples of the weights which are less correlated with each other than if the samples were near to each other in weight-space.
Moreover, the distance of typical samples from the approximate posterior over each layer from the mean is $\sim\sigma\sqrt{D}$, for standard deviation parameter $\sigma$ and the number of parameters in the layer $D$, for the domain of large $D$ typically found in modern neural networks.\footnote{To calculate the distance of typical samples from the mean we imagine an isotropic posterior. Our posteriors are not isotropic, but the pattern is similar.}
We anticipate (and demonstrate in \S\ref{s:understanding}) that the distance between samples from the MFVI approximate posterior makes the gradient estimator of the log-likelihood term of the loss in Equation (\ref{eq:MFVI_unit_prior}) have a large variance, which makes optimization difficult.

\subsection{The Radial BNN Posterior}
A `soap-bubble' arises when, for large $D$, the probability density function over the radius from the mean is sharply peaked at a large distance from the mean (see Figure \ref{fig:pdf_plot}).
Therefore, we pick a probability distribution which cannot have this property.
We can easily write down a probability density function which cannot have a `soap-bubble' by explicitly modelling the radius from the mean.
The hyperspherical coordinate system suits our needs: the first dimension is the radius and the remaining dimensions are angles.
We pick the simplest practical distribution in hyperspherical coordinates with no soap bubble:
\begin{itemize}
    \item In the radial dimension: $r=|\tilde{r}|$ for $\tilde{r} \sim \mathcal{N}(0,1)$.
    \item In the angular dimensions: uniform distribution over the hypersphere---all directions equally likely.
\end{itemize}

\begin{table*}[t]
        \centering
        \resizebox{\textwidth}{!}{\begin{tabular}{llcccccc}
            \toprule
             \multirow{2}{*}{Method} &
             \multirow{2}{*}{Architecture} &
             \multirow{2}{*}{\shortstack{\# \\ Params}} &
             \multirow{2}{*}{\shortstack{Epoch Train \\ Time (m)}} &
             \multicolumn{4}{c}{ROC-AUC for different percent data referred to experts}\\
             &&&&0\% & 10\% & 20\% & 30\%
             \\
             \midrule
             MC-dropout & [Leibig et al., 2017] & $\sim$21M & -& 92.7$\pm$0.3\%&93.8$\pm$0.3\%&94.7$\pm$0.3\%&95.6$\pm$0.3\%\\
             MC-dropout & VGG-16 & $\sim$15M & 5.6 & 93.0$\pm$0.04\% &94.1$\pm$0.05\% &94.5$\pm$0.05\% &95.1$\pm$0.07\%\\
             MFVI & VGG-16* & $\sim$15M & 16.0 & 63.6$\pm$0.13\% &63.5$\pm$0.09\% &63.5$\pm$0.09\% &62.6$\pm$0.10\% \\
             MFVI w/ tweaks & VGG-16* & $\sim$15M & 16.0 & 93.9$\pm$0.04\% &94.4$\pm$0.05\% &95.4$\pm$0.04\% &96.4$\pm$0.05\%\\
             \textbf{Radial BNN} & VGG-16* & $\sim$15M & 16.2 & \textbf{94.3$\pm$0.04\%} &\textbf{95.3$\pm$0.06\%} &\textbf{96.1$\pm$0.06\%} &\textbf{96.8$\pm$0.04\%} \\
             \midrule
             Deep Ensemble & 3xVGG-16 & $\sim$45M & 16.8$\dagger$ & 93.9$\pm$0.04\% & 96.0$\pm$0.05\% & 96.6$\pm$0.04\% &97.2$\pm$0.04\% \\
             \textbf{Radial Ensemble} & 3xVGG-16* & $\sim$45M & 48.6$\dagger$ & \textbf{94.5$\pm$0.05\%} &\textbf{97.9$\pm$0.04\%} &\textbf{98.0$\pm$0.03\%} &\textbf{98.1$\pm$0.03\%}\\
            \bottomrule
        \end{tabular}}
        \caption{Diabetic Retinopathy Prescreening: Our Radial BNN outperforms SOTA MC-dropout and is able to scale to model sizes that MFVI cannot handle without ad-hoc tweaks (see \S\ref{s:tweaks}). Even with tweaks, Radial BNN still outperforms. Deep Ensembles outperform a single Radial BNN at estimating uncertainty, but are worse than an ensemble of Radial BNNs with the same number of parameters. $\pm$ indicates bootstrapped standard error from 100 resamples of the test data. VGG-16* model has fewer channels so that \# of parameters is the same as non-Bayesian model. $\dagger$ 3x single model train time. Could be in parallel.}
        \label{tab:auc}
        \vspace{-3mm}
\end{table*}
A critical property is that it is easy to sample this distribution \textit{in the weight-space coordinate system}---we wish to avoid the expense of explicit coordinate transformations when sampling from the approximate posterior.
Instead of sampling the posterior distribution directly, we use the local reparameterization trick \cite{rezende_stochastic_2014, kingma_auto-encoding_2014}, and sample the noise distribution instead.
This is similar to \citet{graves_practical_2011, blundell_weight_2015} who sample their weights
\begin{equation}
    \textbf{w} \coloneqq \boldsymbol{\mu} + \boldsymbol{\sigma} \odot \boldsymbol{\epsilon}_{\text{MFVI}}, \label{eq:reparameterization}
\end{equation}
where $\boldsymbol{\epsilon}_{\text{MFVI}} \sim \mathcal{N}(0, \mathbf{I})$.
In order to sample from the Radial BNN posterior we make a small modification:
\begin{equation}
    \textbf{w}_{\text{radial}} \coloneqq  \boldsymbol{\mu} + \boldsymbol{\sigma} \odot \frac{\boldsymbol{\epsilon}_{\text{MFVI}}}{\norm{\boldsymbol{\epsilon}_{\text{MFVI}}}} \cdot r ,
\end{equation}
which works because dividing a multi-variate Gaussian random variable by its norm provides samples from a direction uniformly selected from the unit hypersphere \citep{muller_note_1959, marsaglia_choosing_1972}.
As a result, sampling from our posterior is nearly as cheap as sampling from the MFVI posterior.
The only extra steps are to normalize the noise, and multiply by a scalar Gaussian random variable.

\subsection{Evaluating the Objective}
\renewcommand{\wr}{\mathbf{w}^{(r)}}
\newcommand{\wx}{\mathbf{w}^{(x)}}
\newcommand{\er}{\boldsymbol{\epsilon}^{(r)}}
\newcommand{\ex}{\boldsymbol{\epsilon}^{(x)}}
\newcommand{\eri}{\epsilon^{(r)}}
\newcommand{\exi}{\epsilon^{(x)}}
\newcommand{\bmu}{\boldsymbol{\mu}}
\newcommand{\bsi}{\boldsymbol{\sigma}}
\newcommand{\bth}{\boldsymbol{\theta}}
To use the our approximate posterior for variational inference we must be able to estimate the ELBO loss.
The Radial BNN posterior does not change how the expected log-likelihood is estimated, using mini-batches of datapoints and MC integration.

The KL divergence between the approximate posterior and prior can be written:
\begin{align}
    \KLdiv{q(\mathbf{w})}{p(\mathbf{w})} &= \int q(\textbf{w}) \log [q(\textbf{w})] d\textbf{w} \nonumber\\&- \int q(\textbf{w}) \log \big[ p(\textbf{w})\big] d\textbf{w} \nonumber \\ &= \mathcal{L}_{\text{entropy}} - \mathcal{L}_{\text{cross-entropy}}.
\end{align}
We estimate the cross-entropy term using MC integration, just by taking samples from the posterior and averaging their log probability under the prior.
We find that this is low-variance in practice, and is often done for MFVI as well \citep{blundell_weight_2015}.

We can evaluate the entropy of the posterior analytically.
We derive the entropy term in Appendix \ref{a:derivation_of_entropy}:
\begin{align}
    \mathcal{L}_{\text{entropy}} = - \sum_i\log[\sigma_i] + \text{const.}
\end{align}
where $i$ sums over the weights.
This is, up to a constant, the same as when using an ordinary multivariate Gaussian in MFVI.
(For sake of completeness, we also derive the constant terms in the Appendix.)
In Appendix \ref{a:derivation_of_prior}, we also provide a derivation of the cross-entropy loss term in the case where the prior is a Radial BNN.
This is useful in continual learning (see \S\ref{s:vcl}) where we use the posterior from training one model as a prior when training another.

Code implementing Radial BNNs can be found at \url{https://github.com/SebFar/radial_bnn}.

\subsection{Computational Complexity}
Training Radial BNNs has the same computational complexity as MFVI---$\mathcal{O}(D)$, where $D$ is the number of weights in the model.
In contrast, recent non-mean-field extensions to VI like \citet{louizos_structured_2016} and \citet{sun_learning_2017} have higher time complexities.
For example, \citet{louizos_structured_2016} uses a pseudo-data approximation which reduces their complexity to $\mathcal{O}(D + M^3)$ where $M$ is a pseudo-data count.
But even for MNIST, they use $M$ up to 150 which becomes computationally expensive (this is their largest experiment).
\citet{sun_learning_2017} have the same complexity as \citet{louizos_structured_2016}, depending on similar approximations and consider a maximum input dimension of only 16---over 16,000 times smaller than the input dimension of the task we address in \S\ref{s:retinopathy}.
In practice, the comparison between our Radial BNNs and MFVI is even more favorable.
Our method is more robust to hyperparameters, allowing hyperparameters to be selected for training/inference speed to still achieve good accuracy.

\section{EXPERIMENTS}\label{s:experiments}

Our work is focused on large datasets and big models, which is where the most exciting application for deep learning are.
That is where complicated variational inference methods that try to learn weight covariances become intractable, and where the `soap-bubble' pathology emerges.

We address this head-on in \S\ref{s:retinopathy}.
We show that on a large-scale diabetic retinopathy diagnosis image classification task: our radial posterior is \textit{more accurate}, has \textit{better calibrated uncertainty}, and is \textit{more robust} to hyperparameters than MFVI with a multivariate Gaussian and therefore requires significantly fewer iterations and less experimenter time.
In this setting, we have $\sim$260,000 input dimensions and use a model with $\sim$15M parameters.
This is orders of magnitude larger than most other VI work, has been heavily influenced by the experimental settings used to evaluate performance on UCI datasets by \citet{hernandez-lobato_probabilistic_2015} with between 4 and 16 input dimensions and using fewer than 2000 parameters.\footnote{Radial BNNs, like MFVI, do not match the performance of some of the more expensive methods on the UCI datasets. We would not expect it to---our method is specifically designed for models with high-dimensional weight-space, not for the artificial constraints of the experimental settings used on the UCI evaluations. See Appendix \ref{a:uci} for details.}

In \S\ref{s:vcl}, we show that we can use the posterior from variational inference with Radial BNNs as a prior when learning future tasks.
We demonstrate this using a continual learning problem \citep{kirkpatrick_overcoming_2017} on FashionMNIST \citep{xiao_fashion-mnist_2017}.
We show significantly improved performance relative to the MFVI-based Variational Continual Learning (VCL) introduced by \citet{nguyen_variational_2018}.

\subsection{Diabetic Retinopathy Prescreening}\label{s:retinopathy}
\begin{figure}
\vspace{-2mm}
\includegraphics{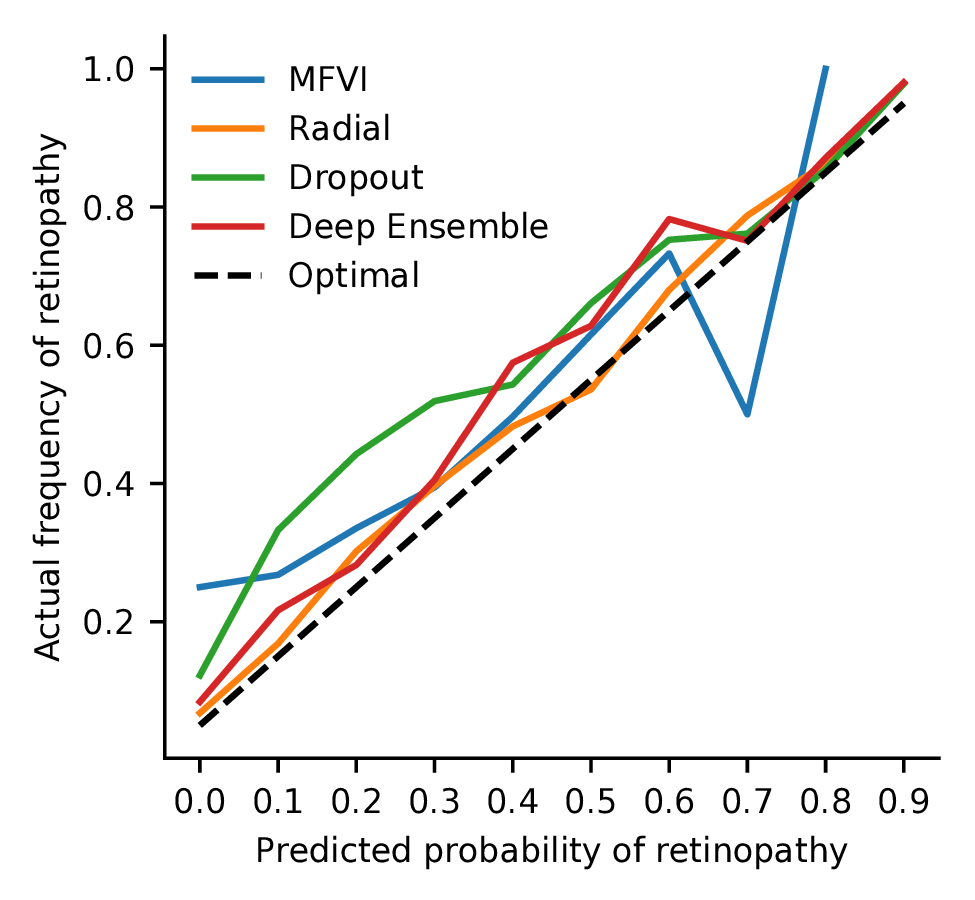}
\vspace{-8mm}
\caption{Radial BNN is almost perfectly calibrated, compared with MC dropout and deep ensembles (overconfident) and ordinary MFVI without ad-hoc tweaks which is not well calibrated. X-axis labels are the lower-bound of each range (e.g., 0.0 is 0.0-0.1).} \label{fig:calibration}
\vspace{-5mm}
\end{figure}
We perform classification on a dataset of `fundus' images taken of the back of retinas in order to diagnose diabetic retinopathy \citep{kaggle_diabetic_2015} building on \citet{leibig_leveraging_2017} and \citet{filos_benchmarking_2019}.
Diabetic retinopathy is graded in five stages, where 0 is healthy and 4 is the worst.
Following \citet{leibig_leveraging_2017}, we distinguish the healthy (classes 0 and 1) from those that require medical observation and attention (2, 3, and 4).
Images (512x512) include left and right eyes separately, which are not considered as a pair by the models, and come from two different camera technologies in many different physical locations.
Model uncertainty is used to identify badly-taken or confusing images which could be used to refer affected patients to experts for more detailed examination.

\subsubsection{Performance and Calibration}
In Table \ref{tab:auc} we compare the classification area under the curve (AUC) of the receiver operating characteristic of predicted classes (higher is better).\footnote{We use AUC because classes are unbalanced (mostly healthy): accuracy gives distorted picture of performance.}
We consider the model performance under different thresholds for referring data to experts.
At 0\%, the model makes predictions about all data.
At 30\%, the 30\% of images about which the model is least confident are referred to experts and do not get scored for the model---the AUC should therefore become higher if the uncertainties are well-calibrated.
We show that our Radial BNN outperforms MFVI by a wide margin, and even outperforms MC dropout.
While the deep ensemble is better at estimating uncertainty than a single Radial BNN, it has three times as many parameters.
An ensemble of Radial BNNs outperforms deep ensembles at all levels of uncertainty.
Radial BNN models trained on this dataset also show empirical calibration that is closer to optimal than other methods (see Figure \ref{fig:calibration}).

The model hyperparameters were all selected individually by Bayesian optimization using ten runs.
Full hyperparameters and search strategy, preprocessing, and architecture are provided in Appendix \ref{a:dr_hypers}.
We include both the original MC dropout results from \citet{leibig_leveraging_2017} as well as our reimplementation using the same model architecture as our Radial BNN model.
The only difference between the MC dropout and Radial BNN/MFVI architectures is that we use more channels for MC dropout, so that the number of parameters is the same in all models.
We estimate the standard error of the AUC using bootstrapping.

\subsubsection{MFVI Tweaks}\label{s:tweaks}
In some cases, researchers have been able to get MFVI to work by applying various ad-hoc tweaks to the training process.
Here, we evaluate the performance of these tweaks and in \S\ref{s:understanding} we explain how the success of these tweaks aligns with our hypothesis that MFVI suffers from a sampling problem which Radial BNNs fix.

One approach to making MFVI work is to pre-train the means of the model using the ordinary log-likelihood loss and to switch partway through training to the ELBO loss, initializing the weight variances at this point with a very small value approximating a deterministic neural network. (E.g., \citep{nguyen_variational_2018} who initialize with a variance of $10^{-6}$ after pre-training the means.)
If one trains to convergence, the weight variances will tend to grow bigger than their tiny initialization, which destroys model performance in MFVI, so one must also employ early stopping.\footnote{Other authors, e.g., \citet{fortunato_noisy_2018}, achieve a similar result just by ignoring the KL to the prior in the loss so that the weight variances tend to shrink to overfit the training data.}
If we perform all these ad-hoc tweaks we are indeed able to get acceptable performance on our diabetic retinopathy dataset (see Table \ref{tab:auc}), though still worse than Radial BNNs.
But the tweaks mean that the learned distribution can certainly not be regarded as an approximate posterior based on optimizing the ELBO.
Moreover, the tweaks amount to approximating a deterministic network.

\subsubsection{Robustness}\label{s:robustness}
\begin{figure}[t]
\centering
\vspace{-2mm}
  \includegraphics[width=\columnwidth]{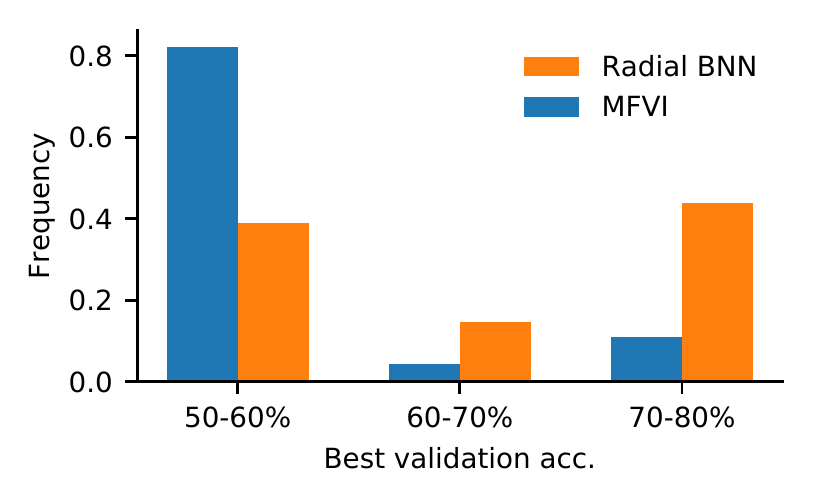}
  \vspace{-8mm}
  \caption{Radial BNN posterior is more robust to hyperparameters on a downsampled version of the retinopathy dataset. Over 80\% of configurations for the MFVI baseline learned almost nothing. 4 times more Radial BNNs had good accuracies than MFVI models.}\label{fig:dr_robust}
\vspace{-2mm}
\end{figure}
The radial posterior was more robust to hyperparameter variation (Figure \ref{fig:dr_robust}).
We assess robustness on a downsampled version of the diabetic retiopathy dataset (256x256) using a smaller model with a similar architecture to VGG-16, but which trained to convergence in about a tenth the time and had only $\sim$1.3M parameters.
We randomly selected 86 different runs from plausible optimizer, learning rate, learning rate decay, batch size, number of variational samples per forward pass, and initial variance.
82\% of hyperparameters tried for the MFVI baseline resulted in barely any improvement over randomly guessing, compared with 39\% for the radial posterior.
44\% of configurations for our radial posterior reached good AUCs, compared with only 11\% for MFVI.
This is despite the fact that we did allow models to pre-train the means using a negative log-likelihood loss for one epoch before begining ELBO training, a common tweak to improve MFVI.

\subsection{Continual Learning}\label{s:vcl}
\begin{figure}[t]
\vspace{-2mm}
  \includegraphics[width=\columnwidth]{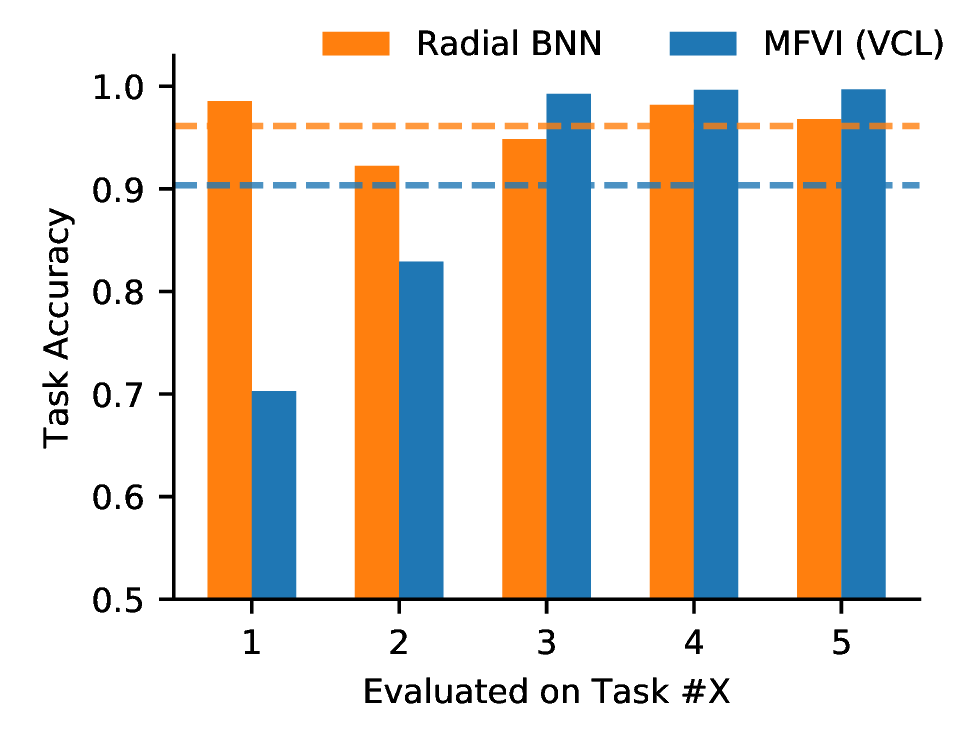}
  \label{fig:multi}
  \vspace{-8mm}
    \caption{Models are trained on the five FashionMNIST tasks in sequence using the posterior from the previous task to remember \textit{all} earlier tasks. Here we show performance of the final model, on all tasks. MFVI (as used in VCL) gradually forgets tasks---the final model's accuracy is worse the older the tasks get. Our Radial BNN preserves information and is still good at the first task. Average accuracy shown by the dotted line.}
    \vspace{-4mm}
\label{fig:vcl}
\end{figure}
\textit{Note: We have been made aware of a missing term in the prior loss for this experiment, see Appendix. We hope to fix this soon.}
Continual learning is a problem setting where a sequence of tasks must be learned separately while a single model is carried from one task to the next but all data are discarded \citep{kirkpatrick_overcoming_2017}.
This is hard because neural networks tend to exhibit `catastrophic forgetting' and lose performance on previous tasks.
A number of authors have proposed prior-focused Bayesian approaches to continual learning in which the posterior at the end of learning a task becomes a prior when learning the next task \citep{kirkpatrick_overcoming_2017, zenke_continual_2017, chaudhry_riemannian_2018, nguyen_variational_2018,farquhar_unifying_2018, ritter_scalable_2018}.
In the case of exact Bayesian updating, this ought to balance the information learned from the datasets of each tasks.
But for approximate methods, we have no such guarantee.
The better the posterior approximation, the better we might expect such prior-focused Bayesian approaches to work.

Variational Continual Learning (VCL), by \citet{nguyen_variational_2018}, applies MFVI to learning the posterior.
Here, we use VCL as a problem setting to evaluate the quality of the posterior.
Note that we do not aim to solve the continual learning problem, but rather to demonstrate the improvement offered by Radial BNNs to the posterior approximation.
A good posterior estimate should work as an effective prior and prevent forgetting.
This setting is particularly relevant to variational inference, as other methods for estimating uncertainty in neural networks (such as Monte-Carlo dropout \citep{gal_dropout_2015} or ensembles \citep{lakshminarayanan_simple_2016}) cannot be straightforwardly used during training as a prior because the posteriors they learn assign zero probability to almost all weight values.

We consider a sequence of five tasks known as Split FashionMNIST \citep{nguyen_variational_2018, farquhar_towards_2018}.
FashionMNIST is a dataset of images of items of clothing or attire (shoes, t-shirts, handbags etc.) \citep{xiao_fashion-mnist_2017}.
The first task is to classify the first two classes of FashionMNIST, then the next two etc.
We examine a multi-headed model \citep{chaudhry_riemannian_2018, farquhar_towards_2018} in order to evaluate the quality of the posterior, although this is a limited version of continual learning.
The models are BNNs with four hidden layers with 200 weights in each ($\sim$250k parameters).
We perform an extensive grid search over hyperparameters.
Full hyperparameters and a more thorough description of the experimental settings, as well as results for the single-headed continual learning setting, are in Appendix \ref{a:vcl_hypers}.

The Radial BNN approximate posterior acts as a better prior, showing that it learns the true posterior better (Figure \ref{fig:vcl}).
Radial BNNs maintain good accuracy on old tasks even after training on all five tasks.
In contrast, the MFVI posterior gets increasingly less accurate on old tasks as training progresses.
The MFVI posterior approximation is not close enough to the true posterior to carry the right information to the next task.
\FloatBarrier
\section{ANALYSING RADIAL BNNs}\label{s:understanding}
Why is it that Radial BNNs offer improved performance relative to MFVI?
In \S\ref{s:method} we observed that the multivariate Gaussian distribution typically used in MFVI features a `soap-bubble'---almost all of the probability mass is clustered at a radius proportional to $\sigma\sqrt{D}$ from the mean in the large $D$ limit (illustrated in Figure \ref{fig:pdf_plot}).
This has two consequences in larger models.
First, unless the weight variances are very small, a typical sample from the posterior has a high $L_2$ distance from the means.
Second, because the mass is distributed uniformly over the hypersphere that the `soap-bubble' clusters around, each sample from the multivariate Gaussian has a high expected $L_2$ distance from every other sample (similarly proportional to $\sigma\sqrt{D}$).
This means that as $\sigma$ and $D$ grow, samples from the posterior are very different from each other, which we might expect to result in high gradient variance.

In contrast, in Radial BNNs the expected distance between samples from the posterior is independent of $D$ for the dimensionality typical of neural networks.
The expected $L_2$ distance between samples from a unit hypersphere rapidly tends to $\sqrt{2}$ as the number of dimensions increases.
Since the radial dimension is also independent of $D$, the expected $L_2$ distance between samples from the Radial BNN is independent of $D$.
This means that, even in large networks, samples from the Radial BNN will tend to be more representative of each other.
As a result, we might expect that the gradient variance is less of a problem.

\begin{figure}
\vspace{-4mm}
\includegraphics{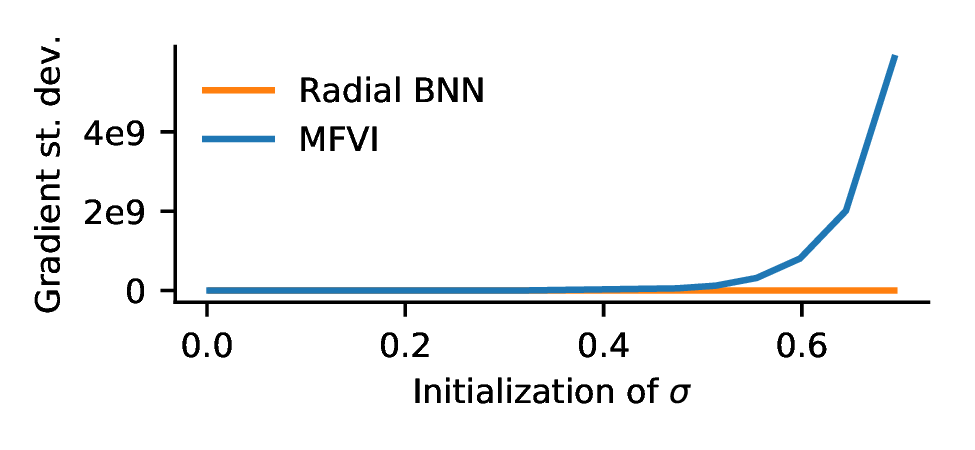}
\vspace{-8mm}
\caption{The variance of gradient estimates in the standard MFVI posterior explodes as the weight variance parameter grows.} \label{fig:gradients}
\vspace{-6mm}
\end{figure}

Indeed, this is exactly what we find.
In Figure \ref{fig:gradients}, we show that for the standard MFVI posterior in a 3x3 conv layer with 512 channels, the variance of initial gradients explodes after the weight standard deviation exceeds roughly $0.3$.
This matters because, for MFVI with a unit Gaussian prior, the KL-divergence term of the loss is minimized by $\sigma_i=1$---well within the region where gradient noise has exploded.

\begin{figure}[t]
\vspace{-4mm}
\includegraphics[width=\columnwidth]{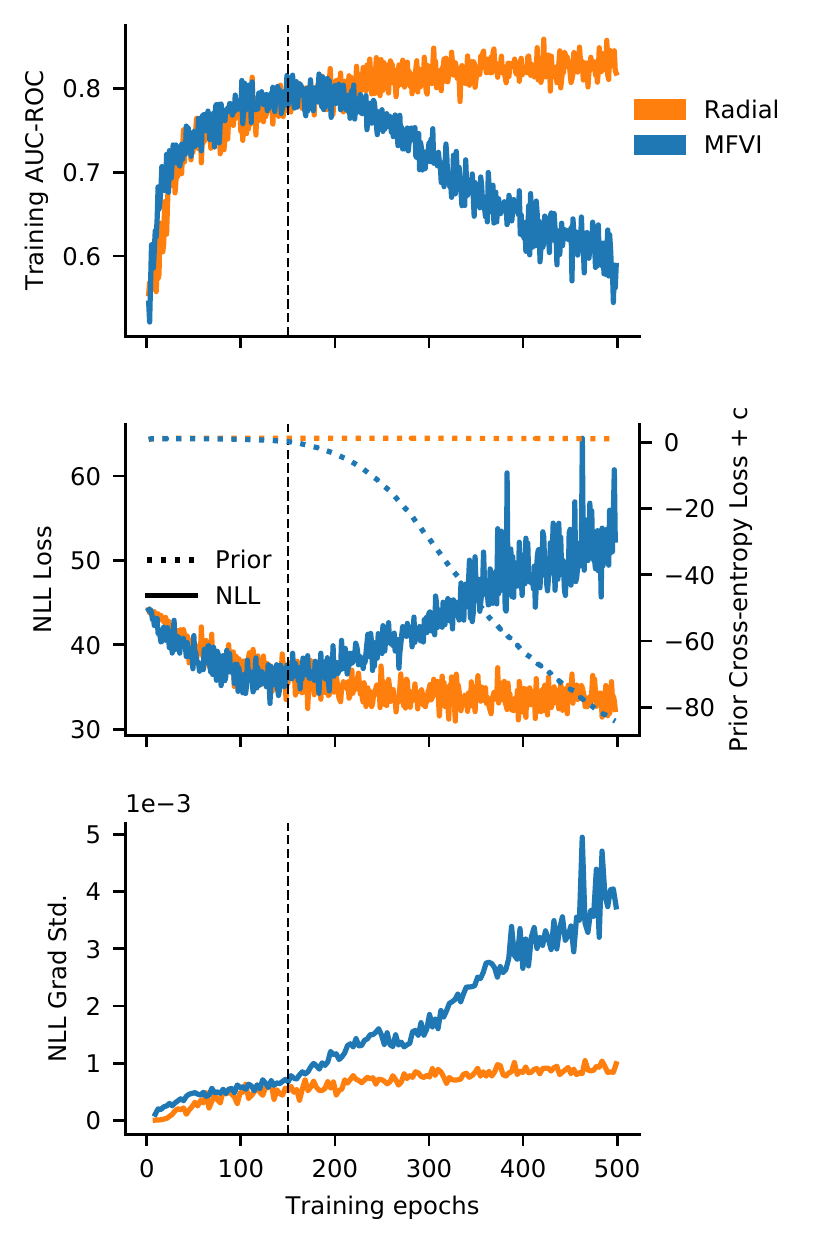}
\vspace{-8mm}
\caption{We can track the deterioration of the MFVI training dynamics. \textbf{Top}: After $\sim$150 epochs (dashed line) \textit{training} set performance degrades for MFVI while Radial continues to improve. \textbf{Middle}: for MFVI, the NLL term of the loss \textit{increases} during training, but the prior cross-entropy term falls faster so the overall loss continues to fall. \textbf{Bottom}: The standard deviation of the NLL gradient estimator grows sharply for MFVI after about 150 epochs. This coincides with the moment where the loss is optimized by minimizing the prior cross-entropy and sacrificing the NLL.} \label{fig:training_loss}
\vspace{-6mm}
\end{figure}

We can track this effect as it kicks in during training.
In Figure \ref{fig:training_loss} we show a sample training run on the down-sampled version of the diabetic retinopathy dataset using an MFVI and Radial BNN with the same hyperparameters.
Pathologically, the \textit{training} accuracy falls for MFVI after about 150 epochs (top graph).
The critical moment corresponds to the point where the training process begins to optimize the prior cross-entropy term of the loss, sacrificing the negative log-likelihood term (middle graph).
We can further show that this corresponds to the point where the standard deviation of the negative log-likelihood term of the gradient begins to sharply increase for MFVI.
Meanwhile, the prior cross-entropy term is computed analytically, so its variance does not grow as the values of $\sigma$ increase during training from their tiny initializations.

That is, MFVI fails because the high variance of the negative log-likelihood (NLL) term of the loss causes the optimizer to improve the cross-entropy term at the expense of the NLL term.
For Radial BNNs, however, the NLL gradient variance stays low throughout training.

In Appendix \ref{a:further_grad_exp} we offer further analysis of the failure of MFVI demonstrating that a biased but low-variance estimator of the gradient (using a truncated posterior approximation) improves training in MFVI.

\FloatBarrier
\subsection{Avoiding the Pathology in MFVI}
Based on this analysis, we can see why Radial BNNs fix a sampling problem in MFVI.
But this also helps explain why the ad-hoc tweaks which researchers have been using for MFVI have been successful.
These tweaks chiefly serve to keep the weight variance low.
Researchers initialize with small variances \citep{blundell_weight_2015, fortunato_bayesian_2017, nguyen_variational_2018}.
Sometimes they adapt the loss function to remove or reduce the weight of the KL-divergence term, which reduces the pressure on weight variances to grow \citep{fortunato_noisy_2018}.
Other times researchers pretrain the means with just the NLL loss, which makes it possible to stop training after relatively little training on the ELBO loss, which stops the variances from growing too much \citep{nguyen_variational_2018}.
Another approach, which we have not seen tried, would be to use a very tight prior, effectively enforcing the desire to have a basically deterministic network (a prior inversely proportional to $\sqrt{D}$ would balance the `soap-bubble' variance).
However, this sort of very tight prior is not compatible with the use of data-dependent priors in sequential learning.

For most of these tweaks, the resulting network is not fully optimizing the ELBO.
This does not necessarily make the resulting network useless---after all, the ELBO is only a bound on the actual model evidence, and other methods like Deep Ensembles work surprisingly well despite not necessarily estimating the model posterior at all.
However, if we have a theoretically principled way to fix our sampling problems without resorting to ad-hoc tweaks, then we should prefer that.
Radial BNNs offer exactly that theoretically principled fix.

\section{Discussion}
Bayesian neural networks need to scale to large models in order to reach their full potential.
Until now, researchers who want BNNs at scale needed to accept a posterior distribution which has zero probability mass almost everywhere---an implausible and problematic assumption.
At the same time, MFVI requires increasingly demanding ad-hoc tweaks in order to work in anything but small models.
We show why MFVI faces a serious gradient estimation problem which gets worse in high dimensions.
Based on this motivation, we introduce Radial BNNs.
This alternative variational inference posterior approximation is simple to implement, computationally fast, robust to hyperparameters, and scales to large models.
Radial BNNs outperform other efficient BNN methods, and have the potential to craft data-dependent priors for use in applications like continual learning.

\FloatBarrier
\section*{Acknowledgements}
We would like to especially acknowledge Alexander Lvovsky for spotting an error in an early version of a proof and Lewis Smith for useful conversations and suggestion of the experiment in Appendix \ref{a:further_grad_exp}.
We would also like to thank Milad Alizadeh, Joost van Amersfoort, Gregory Farquhar, Angelos Filos, and Andreas Kirsch for valuable discussions and/or comments on drafts.
In addition, we gratefully thank the Alan Turing Institute and Google for their donation of computing resources.
Lastly, we thank the EPSRC for their support of Sebastian Farquhar through the Centre for Doctoral Training in Cyber Security at the University of Oxford.

\bibliography{references}
\bibliographystyle{plainnat}

\newpage
\appendix
\appendixpage
\section{Understanding the Soap Bubble}\label{a:soap_bubble}

The emergence of a `soap-bubble' is a well-known property in multi-variate Gaussian distributions as the number of dimensions increases (e.g., see \citet{bishop_introduction_2006}).
The observation is that, even though the highest probability density is near the mean, because there is just \textit{so much more volume} further from the mean in high-dimensional spaces it ends up being the case that most of the probability mass is far from the mean.

One way to understand this is to examine the probability density function of the multivariate Gaussian along its radius.
Consider a $D$-dimensional isotropic Gaussian.
We examine a thin shell with thickness $\eta$, which tends to zero, at distance $r$ between a sampled point, $\mathbf{w}$, and the mean of the multivariate Gaussian, $\boldsymbol{\mu}$.
The probability density function over the radius is given by:
\begin{equation}
\begin{split}
    \lim_{\eta\to0} p(r - \eta < \norm{\mathbf{w} - \boldsymbol{\mu}} < r + \eta) \\= \frac{S_D}{(2\pi\sigma^2)^{D/2}} \cdot r^{D-1} \cdot e^{-\frac{r^2}{2\sigma^2}}
    \label{eq:soap_bubble}
\end{split}
\end{equation}

where $S_D$ is the surface area of a hypersphere in a $D$-dimensional space.

The first term of the product is just a normalizing constant ($S_D$ is the surface area of a D-dimensional hypersphere).

\begin{figure}
    \centering
    \includegraphics[width=\columnwidth]{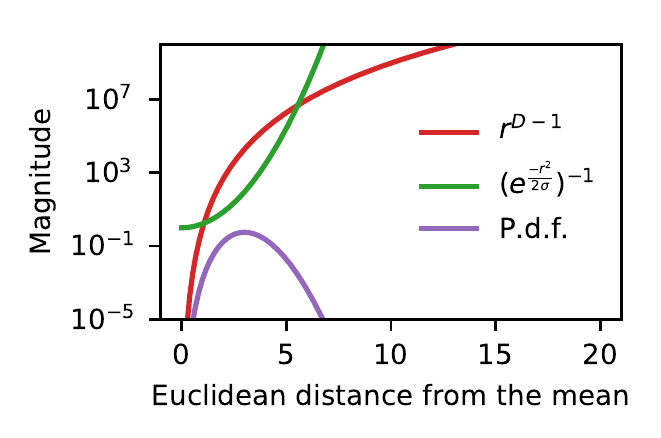}
    \caption{We can understand the `soap-bubble' by looking at components of the p.d.f. in eq. \ref{eq:soap_bubble}. For small $r$ the volume term (red) dominates and the normalized p.d.f. is very small. For big $r$ the Gaussian density term (inverse shown in green) dominates and the p.d.f. is small again. Almost all the probability mass is in the intermediate region where neither term dominates: the `soap bubble'. The intermediate region becomes narrower and further from the mean as $D$ is bigger. Here we show $D=10$.}
    \label{fig:soap_bubble}
\end{figure}

The second term, $r^{D-1}$, reflects the growing \textit{volume} in shells away from the origin.
In the region where $r$ is small, and for the large $D$ found in BNNs with many parameters, this term (red in figure \ref{fig:soap_bubble}) dominates and drives the probability density towards zero.

The exponential term $e^{-\frac{r^2}{2\sigma^2}}$ reflects the Gaussian density (inverse shown in green in figure \ref{fig:soap_bubble}).
For larger $r$ the exponential term becomes very small and drives the probability density towards zero.
Almost all the probability mass is in the `soap-bubble' in the region where neither term becomes very small.
We consider the isotropic case here for simplicity, but the non-isotropic Gaussian has a similar `soap-bubble'.\footnote{\citet{oh_BOCK_2018} consider `soap-bubbles' in Bayesian optimization.
But it has not been considered for MFVI.}

\subsection{Radial Approximate Posterior Over Each Weight}
In order to achieve an approximate posterior distribution which does not have a soap bubble, we must use a lighter tailed distribution over each weight than a Gaussian would.
\begin{figure}
    \centering
    \includegraphics[width=\columnwidth]{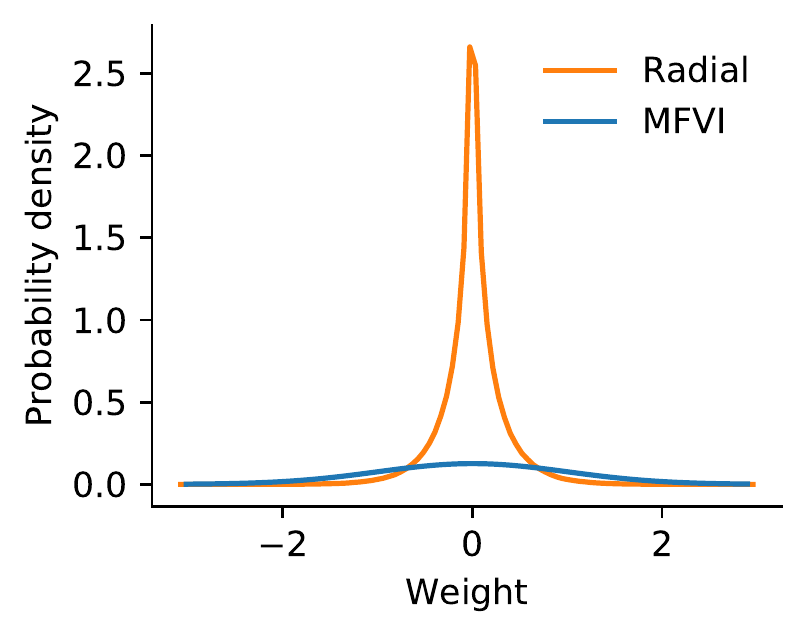}
    \caption{One-dimensional marginal distribution for a single weight. For each single weight, the Radial approximate posterior is lighter-tailed than the MFVI Gaussian, and more so the larger the layer. Here, we show a 10-dimensional layer.}
    \label{fig:marginal}
\end{figure}

In Figure \ref{fig:marginal} we show the distribution for a single weight from a Radial BNN layer with 10 weights.
It is much more sharply peaked than for a typical multivariate Gaussian and has lighter tails.
As a result, each individual weight in a sample from a Radial BNN is much more similar to other samples of that weight.
From the perspective of viewing the model function as a sample from the entire weight-space, however, the Radial BNN distribution is more attractive, because it does not display the `soap-bubble' pathology.

\section{Derivation of the Entropy Term of the KL-divergence}\label{a:derivation_of_entropy}
In this section, we show that the component of KL-divergence term of the loss which is the entropy of the posterior distribution over the weights $q(\wx)$ can be estimated as:

\begin{align}
    \mathcal{L}_{\text{entropy}} \coloneqq& \int q(\wx) \log [q(\wx)] d\wx \label{eq:entropy}\\
    =&- \sum_i\log[\sigma_i^{(x)}] + \text{const}
\end{align}

where $i$ is an index over the weights of the model.

Throughout this section we use a superscript indicates the basis---an $(x)$ means we are in the Cartesian coordinate system tied to the weight-space while $(r)$ is the hyperspherical coordinate system (the letter is the canonical `first' coordinate of that coordinate system).

We begin by applying the reparameterization trick \citep{kingma_auto-encoding_2014, rezende_stochastic_2014}.
Following the auxiliary variable formulation of \citet{gal_uncertainty_2016}, we express the probability density function of $q(\wx)$ with an auxiliary variable.
\begin{align}
    q(\wx) &= \int q(\wx, \er)d\er\\
    &= \int q(\wx|\er)q(\er)d\er\\
    &= \int \delta(\wx - g(\mu, \sigma, \er)q(\er)d\er. \label{auxilliary}
\end{align}

In equation (\ref{auxilliary}), we have used a reparameterization trick transformation:
\begin{align}
    g(\mu, \sigma, \er) = \bmu + \bsi \odot \mathbf{T}_{rx}(\er)
\end{align}
where $\bmu$ and $\bsi$ are parameters of the model and where $\mathbf{T}_{rx}$ is the standard transformation from hyperspherical into Cartesian coordinates.

Substituting equation (\ref{auxilliary}) into the definition of the entropy loss term in equation (\ref{eq:entropy}), and applying the definition of the Kronecker delta we can eliminate dependence on $\wx$:

\begin{align}
\mathcal{L}_{\text{entropy}} &= \int q(\wx) \log [q(\wx)] d\wx\\    
    &= \int \bigg(\int \delta(\wx - g(\bmu, \bsi, \er) \nonumber \\
    &\qquad\qquad q(\er)d\er\bigg) \log [q(\wx)] d\wx\\
    &= \int q(\er) \log [q(g(\bmu, \bsi, \er))] d\er.
\end{align}

Then, we perform a coordinate transformation from $g(\bmu, \bsi, \er)$ to $\er$ using the Jacobian of the transformation and simplify.
\begin{align}
    &= \int q(\er) \log \Bigg[q(\er) \bigg|\frac{\partial g(\bmu, \bsi, \er)}{\partial \er}\bigg|^{-1}\Bigg]d\er\\
    &= \int q(\er) \log \Bigg[q(\er) \bigg|\prod_i \sigma_i^{(x)} \frac{\partial \exi_i}{\partial \eri_j}\bigg|^{-1}\Bigg]d\er\\
    &= \int q(\er) \log \Bigg[q(\er) \bigg|\text{diag}(\bsi) \frac{\partial \exi_i}{\partial \eri_j}\bigg|^{-1}\Bigg]d\er\\
    &= \int q(\er) \log \Bigg[\frac{q(\er)}{\prod_i \sigma_i^{(x)}} \bigg|\frac{\partial \exi_i}{\partial \eri_j}\bigg|^{-1}\Bigg]d\er \label{eq:unworkedout_entropy}
\end{align}
In the last line we have used the fact that $\forall{i}:$ $\sigma_i^{(x)} \geq 0$ allowing us to pull the determinant of this diagonal matrix out.

$\bigg|\frac{\partial \exi_i}{\partial \eri_j}\bigg|$ is the determinant of the Jacobian for the transformation from Cartesian to hyperspherical coordinates for which we use the result by \citet{muleshkov_easy_2016}:
\begin{align}
    \bigg|\frac{\partial \ex_i}{\partial \er_j}\bigg| &= \text{abs}\Bigg(\big(-1)^{D-1}\big(\epsilon_0^{(r)}\big)^{D-1} \prod_{i = 2}^{D}\big(\sin(\epsilon_i^{(r)})\big)^{i-1}\Bigg).
\end{align}

We know that $\epsilon_0^{(r)} \geq 0$ because the radial dimension in hyperspherical coordinates can be assumed positive without loss of generality.
We also know $0 \leq \epsilon_i^{(r)} \leq \pi$ for $2 \leq i \leq D$ for the hyperspherical coordinate system.
So we can simplify the signs:
\begin{align}
    = \big(\epsilon_0^{(r)}\big)^{D-1} \prod_{i = 2}^{D}\big(\sin(\epsilon_i^{(r)})\big)^{i-1}.\label{eq:coord_conversion}
\end{align}

Therefore, plugging equation (\ref{eq:coord_conversion}) into (\ref{eq:unworkedout_entropy}):
\begin{align}
    \mathcal{L}_{\text{entropy}} &=  \int q(\er) \log \Bigg[\frac{q(\er)}{\text{abs}(\prod_i \sigma_i^{(x)})} \bigg|\frac{\partial \exi_i}{\partial \eri_j}\bigg|^{-1}\Bigg]d\er\\
    &=  \int q(\er) \log [q(\er)] \nonumber \\ &\qquad- \log[\text{abs}(\prod_i \sigma_i^{(x)})] \nonumber \\ &\qquad -\log \bigg[ \big(\epsilon_0^{(r)}\big)^{D-1} \prod_{i = 2}^{D}\big(\sin(\epsilon_i^{(r)})\big)^{i-1}\bigg] d\er.\label{eq:3_terms}
\end{align}

Very simply, we can observe that only the middle term depends on the parameters and we must therefore only compute this term in order to compute gradients.
For sake of completeness, we address the other integrals below, in case one wants to have the full value of the loss (though since it is a lower bound in any case, the full value is not very useful).

The probability density function of the noise variable is separable into independent distributions. The distribution of $\epsilon^{(r)}_0$ is a unit Gaussian. The angular dimensions are distributed so that sampling is uniform over the hypersphere. However, this does \textbf{not} mean that the distribution over each angle is uniform, as this would lead to bunching near the n-dimensional generalization of the poles. (Intuitively, there is more surface area per unit of angle near the equator, as is familiar from cartography.) Instead, we use the fact that the area element over the hypersphere is:
\begin{align}
    dA = d\eri_{D}\prod_{i=1}^{D-1}\sin(\eri_{i})^{D-i}d\eri_{i}
\end{align}
where we remember that $\eri_{D}$ is between $-\pi$ and $\pi$, and the rest of the angular elements of $\er$ are between $0$ and $\pi$. The resulting probability density function is:
\begin{align}
    q(\boldsymbol{\epsilon}^{(r)}) &= \prod_{i=0}^{D} q(\epsilon_i^{(r)})\\
    &= \frac{1}{\sqrt{2\pi}}e^{-\frac{\epsilon_0^2}{2}} \cdot \prod_{i=1}^{D-1}\sin(\eri_{i})^{D-i}. \label{eq:pdf}
\end{align}

As a result, all three of the terms in equation (\ref{eq:3_terms}) are analytically tractable. Inserting the probability density function from equation (\ref{eq:pdf}) into the first term of the loss, splitting up the product inside the logarithm, and separating independent terms we get:

\begin{align}
    &\int q(\er) \log [q(\er)] d\er\nonumber \\
    &= \int_{0}^{\infty}d\eri_0 \int_{-\pi}^{\pi}d\eri_D \int_0^{\pi}\prod_{i=1}^{D-1}d\eri_{i} \nonumber\\
    & \qquad\cdot \frac{1}{\sqrt{2\pi}}e^{-\frac{\epsilon_0^2}{2}} \cdot \prod_{i=1}^{D-1}\sin(\eri_{i})^{D-i} \nonumber\\
    & \qquad \cdot \log\Big[\frac{1}{\sqrt{2\pi}}e^{-\frac{\epsilon_0^2}{2}} \cdot \prod_{i=1}^{D-1}\sin(\eri_{i})^{D-i}\Big]\\
    &= \int_{0}^{\infty}d\eri_0 \int_{-\pi}^{\pi}d\eri_D \int_0^{\pi}\prod_{i=1}^{D-1}d\eri_{i} \nonumber\\
    & \qquad \cdot \frac{1}{\sqrt{2\pi}}e^{-\frac{\epsilon_0^2}{2}} \cdot \prod_{i=1}^{D-1}\sin(\eri_{i})^{D-i} \nonumber\\
    & \qquad \cdot \log\Big[\frac{1}{\sqrt{2\pi}}e^{-\frac{\epsilon_0^2}{2}}\Big] + \sum_{i=1}^{D-1}\log\big[\sin(\eri_{i})^{D-i}\big]\\
    &= \int_{0}^{\infty}d\eri_0 \frac{1}{\sqrt{2\pi}}e^{-\frac{\epsilon_0^2}{2}} \log\Big[\frac{1}{\sqrt{2\pi}}e^{-\frac{\epsilon_0^2}{2}}\Big] \nonumber\\
    & \qquad+ \int_{-\pi}^{\pi}d\eri_D \nonumber\\
    & \qquad+ \sum_{i=1}^{D-1} \int_0^{\pi}d\eri_{i} \sin(\eri_{i})^{D-i} \log\big[\sin(\eri_{i})^{D-i}\big]\label{eq:intractable}\\
    &= -\frac{\log[2\pi] + 1}{4} + 2\pi \nonumber \\
    & \qquad+ \sum_{i=1}^{D-1} \int_0^{\pi}d\eri_{i} \sin(\eri_{i})^{D-i} \log\big[\sin(\eri_{i})^{D-i}\big]
\end{align}

We can simplify the first two of the terms in equation \ref{eq:intractable}, but the third is difficult to solve in general (though tractable for any specific $D$).
Regardless, this term is constant and therefore not needed for our optimization.

Inserting the probability density function from equation (\ref{eq:pdf}) into the second term of the loss in equation (\ref{eq:3_terms}) we get
\begin{align}
    \int q(\er) \log[\prod_i \sigma_i^{(x)}] d\er &= \log[\prod_i \sigma_i^{(x)}]\\
    &= \sum_i\log[\sigma_i^{(x)}].
\end{align}
This second term is identical to the entropy of the multivariate Gaussian variational posterior typically used in MFVI.

And for the third term we begin by expanding the logarithm and simplifying:
\begin{align}
    & \int q(\er) \log \bigg[ \big(\epsilon_0^{(r)}\big)^{D-1} \prod_{i = 2}^{D}\big(\sin(\epsilon_i^{(r)})\big)^{i-1}\bigg] d\er\\
    &= (D-1)\int_0^\infty q(\epsilon_0^{(r)}) \log [ \epsilon_0^{(r)}]d\epsilon_0^{(r)}\nonumber \\
    &\qquad+ \sum_{i = 2}^{D} \frac{(i - 1)}{\pi}\int_0^\pi\log \bigg[\sin(\epsilon_i^{(r)})\bigg] d\epsilon_i^{(0)}\\
\end{align}
and then inserting the p.d.f. from equation (\ref{eq:pdf}) and solving analytically tractable integrals:
\begin{align}
    &= \frac{(D-1)}{\sqrt{2\pi}}\int_0^\infty e^{-\frac{\big(\epsilon_0^{(r)}\big)^2}{2}} \log [ \epsilon_0^{(r)}]d\epsilon_0^{(r)} \nonumber \\
    & \qquad+ \sum_{i = 2}^{D} \frac{(i - 1)}{\pi} \cdot -\pi\log[2]\\
    &= \frac{(D-1)}{\sqrt{2\pi}}\cdot -\frac{1}{2}\sqrt{\frac{\pi}{2}}(\gamma + \log[2]) - \sum_{i = 2}^{D} (i - 1) \cdot \log[2]\\
    &= -\frac{(D-1)}{4}\cdot(\gamma + \log[2]) - \frac{(D-1)(D-2)}{2} \log[2]\\
    &= -\frac{(D-1)\gamma}{4} - \frac{(D-1)(2D-3)}{4} \log[2].\\
\end{align}
where $\gamma$ is the Euler-Mascheroni constant.
This is, again, constant and may be neglected for optimization.

As a result, we can minimize the entropy term of the loss simply by finding

\begin{align}
    \mathcal{L}_{\text{entropy}} = - \sum_i\log[\sigma_i^{(x)}] + \text{const}
\end{align}

\section{Setting a Radial Prior}\label{a:derivation_of_prior}
\textit{Authors' Note: we have been made aware of a mistake in the derivation of the cross-entropy term in the case of a Radial prior.
In particular, we are missing a term for the Jacobian of the transformation between the $\epsilon$-space of the prior and the $\epsilon$-space of the approximate posterior.
We hope to fix this soon, but in the mean-time are adding this note for readers.
We include the previous derivation below, because it is the one used in the experiments in the main paper, but note that it is missing an unknown term.}

In most of our experiments, we use a typical multivariate Gaussian unit prior in order to ensure comparability with prior work.
However, in some settings, such as the Variational Continual Learning setting, it is useful to use the radial posterior as a prior.
We begin similarly to the previous derivation, with all unchanged expect that we are estimating
\begin{align}
    \mathcal{L}_{\text{cross-entropy}} &= \int q(\wx) \log [p(\wx)] d\wx.
\end{align}
The derivation proceeds similarly until equation (\ref{eq:3_terms}), and the second and third terms are identical except the second term taking a product over elements of $\bsi^{(x)}_{(\text{prior})}$ of the prior, not the posterior.

Evaluating the gradient of the log probability density function of the prior depends only on the radial term, since the distribution is uniform in all angular dimensions.
We therefore find
\begin{align}
\int q(\er) \log [p\Big(\frac{\wx - \bmu^{(x)}_{(\text{prior})}}{\bsi^{(x)}_{(\text{prior})}}\Big)] d\er.
\end{align}
Rather than solve the integral, we can estimate this as a Monte Carlo approximation:
\begin{align}
\approx \frac{1}{N}\sum_{i=1}^{N} -\frac{1}{2} \left\lVert\frac{\wx - \bmu^{(x)}_{(\text{prior})}} {\bsi^{(x)}_{(\text{prior})}}\right\rVert^2.
\end{align}
By adding the three terms we estimate the cross-entropy term of the ELBO loss function.

\section{Experimental Settings}\label{a:experimental_settings}
\subsection{Diabetic Retinopathy Settings}\label{a:dr_hypers}
\textit{Authors' Note: The version of the diabetic retinopathy dataset used in this paper (as described below) is slightly different from the one used in the latest release of the Diabetic Retinopathy benchmark (as of May 2021).
Radial BNNs have been included in a \href{https://github.com/google/uncertainty-baselines/tree/master/baselines/diabetic_retinopathy_detection}{more recent benchmarking effort} in which they do not perform quite as well as we found originally.}

The diabetic retinopathy data are publicly available at \url{https://www.kaggle.com/c/diabetic-retinopathy-detection/data}.
We augment and preprocess them similarly to \citet{leibig_leveraging_2017}.
The images for our main experiments in \S\ref{s:retinopathy} are downsampled to 512x512 while the smaller robustness experiment in \S\ref{s:robustness} uses images downsampled to 256x256
We randomly flip horizontally and vertically.
Then randomly rotate 180 degrees in either direction.
Then we pad by between 0 and 5\% of the width and height and randomly crop back down to the intended size.
We then randomly crop to between 90\% and 110\% of the image size, padding with zeros if needed.
We finally resize again to the intended size and normalize the means and standard deviations of each channel separately based on the training set means and standard deviations.
The training set has 44,594 RGB images.
There are 7,026 validation and 10,000 test images. 

The smaller model used for robustness experiments is loosely inspired by VGG-16, with only 16 channels, except that it is a Bayesian neural network with mean and standard deviations for each weight, and that instead of fully connected networks at the end it uses a concatenated global mean and average pool.
The larger model used in the main experiments is VGG-16 but with the concatenated global mean and average pool instead of fully connected layers as above.
The only difference is that we use only 46 channels, rather than 64 channels as in VGG-16, because the BNN has twice as many parameters as a similarly sized deterministic network and we wanted to compare models with the same number of parameters.
For the dropout model we use VGG-16 with the full 64 channels, and similarly for each of the models in the deep ensemble.
The prior for training MFVI and Radial BNNs was a unit multivariate Gaussian.
(We also tried using the scale mixture prior used in \citet{blundell_weight_2015} and found it made no difference.)
Instead of optimizing $\sigma$ directly we in fact optimize $\rho$ such that $\sigma = \log(1 + e^{\rho})$ which guarantees that $\sigma$ is always positive.
In some cases, as described in the paper, the first epoch only trained the means and uses a NLL loss function.
This helps the optimization, but in principle can still allow the variances to train fully if early stopping is not employed (unlike reweighting the KL-divergence).
Thereafter we trained using the full ELBO loss over all parameters.
Unlike some prior work using MFVI, we have not downweighted the KL-divergence during training.

For the larger models, we searched for hyperparameters using Bayesian optimization.
We searched between 0 and -10 as the initial value of $\rho$ (equivalent to $\sigma$ values of $\log(2)$ and $~2\cdot10^{-9}$).
For the learning rate we considered $10^{-3}$ to $10^{-5}$ using Adam with a batch size of 16.
Otherwise, hyperparameters we based on exploration from the smaller model.

We then computed the test scores using a Monte Carlo estimate from averaging 16 samples from the variational distribution.
We estimate the model's uncertainty about a datapoint using the mutual information between the posterior's parameters and the prediction on a datapoint.
This estimate is used to rank the datapoints in order of confidence and compute the model's accuracy under the assumption of referring increasingly many points to medical experts.

For the smaller models, we performed an extensive random hyperparameter search.
We tested each configuration with both MFVI and Radial BNNs.
We tested each configuration for both an SGD optimizer and Amsgrad.
When training with SGD we used Nesterov momentum 0.9 and uniformly sampled from 0.01, 0.001 and 0.0001 as learning rates, with a learning rate decay each epoch of either 1.0 (no decay), 0.98 or 0.96.
When training with Amsgrad we uniformly sampled from learning rates of 0.001, 0.0001, and 0.00001 and did not use decay.
We uniformly selected batch sizes from 16, 32, 64, 128, and 256.
We uniformly selected the number of variational distribution samples used to estimate the loss from 1, 2, and 4.
However, because we discarded all runs where there was insufficient graphics memory, we were only able to test up to 64x4 or 256x1 and batch sizes above 64 were proportionately less likely to appear in the final results.
We selected the initial variance from $\rho$ values of -6, -4, -2, or 0.
We also tried reducing the number of convolutional channels by a factor of $5/8$ or $3/8$ and found that this did not seem to improve performance.
We ran our hyperparameter search runs for 150 epochs.
We selected the best hyperparameter configurations based on the best validation accuracy at any point during the training.
We trained the models for 500 epochs but selected the models saved from 300 epochs as all models had started to overfit by the end of training.
For MFVI, this was using the SGD optimizer with learning rate 0.001, decay rate 0.98 every epoch, batch size 16, 4 variational samples for estimating the loss during training and $\rho$ of -6.
This outperformed the others by a significant margin.
Using our code on a V100 GPU with 8 vCPUs and an SSD this took slightly over 13 hours to train each model.
For the radial posterior, this was the Adam optimizer with learning rate 0.0001, batch size 64, 1 variational sample for estimating the loss during training and a $\rho$ of -6.
Using our code on the same GPU, this took slightly over 3h to run.
However, for the radial posterior there were very many other configurations with similar validation accuracies (one of the advantages of the posterior).

For the experiment shown in Figure \ref{fig:training_loss}, we have selected slightly different hyperparameters in order to train more quickly.
For both models, we use Adam with learning rate 0.0001 and train for 500 epochs.
The models have $5/8$ the number of channels of VGG-16.
The models are trained with batch size 64 and 4 variational samples to estimate the loss and its standard deviation.

\subsection{Variational Continual Learning Settings}\label{a:vcl_hypers}
We build on the code provided by \citet{nguyen_variational_2018} at \url{https://github.com/nvcuong/variational-continual-learning} adapted for FashionMNIST.
The FashionMNIST dataset was downloaded using pytorch's built in vision datasets.
The data were normalized by subtracting the training set mean and dividing by the training set standard deviation.

The classes are ordered in the conventional order.
The model is initialized randomly---without pretraining the means (unlike \citet{nguyen_variational_2018}).
The model is then trained on the first two classes.
The weights are carried over to the next task and set as a prior, while the model is trained on the next two classes, and so on.
Note that we perform the tasks in a multi-headed way---each task has its own output head.
This may not be an ideal exemplar of the continual learning problem \citep{chaudhry_riemannian_2018, farquhar_towards_2018} but it forms an effective test of the posterior.
We do \textit{not} use coresets, unlike \citet{nguyen_variational_2018}, as this would \textit{not} form an effective test of the quality of the posterior.

Models are Bayesian MLPs with four hidden layers with 200 units in each.
The prior for training was a unit multivariate Gaussian.
Instead of optimizing $\sigma$ directly we in fact optimize $\rho$ such that $\sigma = \log(1 + e^{\rho})$ which guarantees that $\sigma$ is always positive.
Models are optimized using Amsgrad \citep{reddi_convergence_2018} with learning rate 0.001 with shuffling and discarding final incomplete batches each epoch.
We perform a grid search over the number of epochs each task is trained over (3, 5, 10, 15, 20, 60, 120) and batch sizes (1024, 2048, 10000).
We used 90\% of the standard training dataset (54000 points) as a training dataset, with 10\% (6000 points) withheld as a validation dataset.
We initialize $\rho$ to $-6$ and use the initialization by \citet{he_deep_2016} for the means.
The radial posterior would work with a much larger $\rho$, but we wanted to use comparable initializations for each.
We optimized for average validation accuracy over all models on the final task.
We used the standard 10000 points as a test dataset.
The best configuration for the MFVI posterior was found to be 60 epochs of batch size 1024 (note that this differs from the 120 epochs of batch size 12000 reported in \citet{nguyen_variational_2018} perhaps because they pretrain the means).
The best configuration for the radial posterior was found to be 20 epochs of batch size 1024.
We report the individual accuracies for each head on the test dataset.

\subsection{Single-headed FashionMNIST continual learning}\label{a:single_headed}
Previous authors have noted that for \textit{continual learning} the single-headed environment---where the model has a single output head shared over all tasks and must therefore identify the task as well as the output based on the input---is a much harder task, possibly more reflective of continual learning \citep{chaudhry_riemannian_2018, farquhar_towards_2018}.
While the multi-headed setting suffices to demonstrate improvement to the posterior, we offer some results for the single-headed setting here in the appendix for the interest of continual learning specialists, though we do not find that our posterior solves the problem.

We perform a similar grid search as before, selecting the hyperparameters that offer the highest average validation set accuracy for the final model over all five tasks.
Note that in our grid search each task gets the same hyperparameters, reflecting the idea that the task distribution is not known in advance.

\begin{figure}
\vspace{-1mm}
  \includegraphics[width=\linewidth]{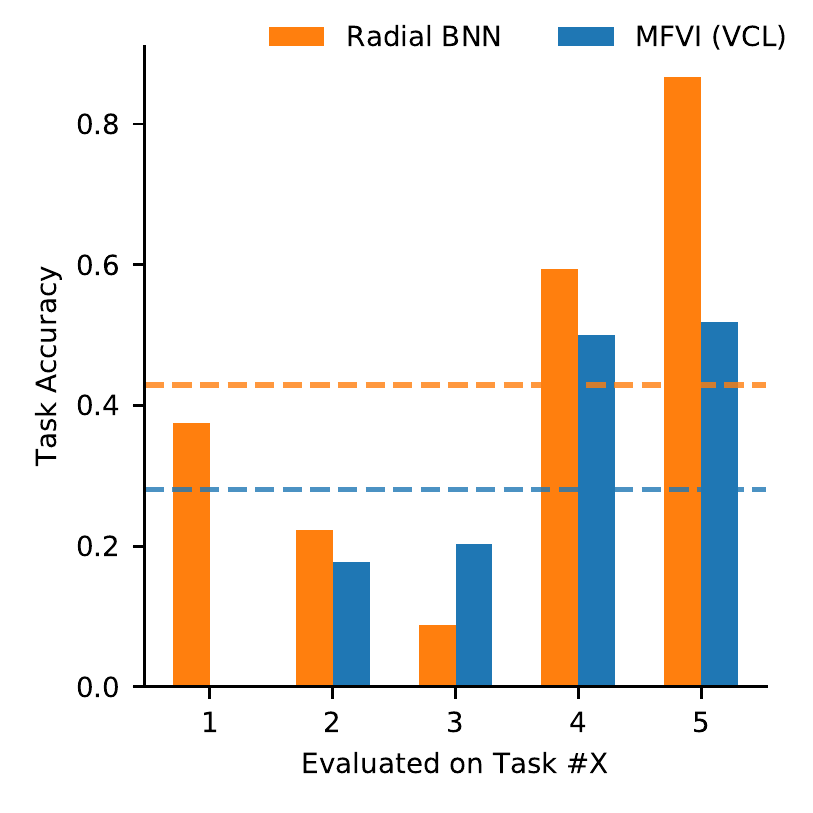}
  \vspace{-8mm}
    \caption{FashionMNIST. In the single-headed setting, where all training and testing ignores the task label, the situation is more difficult. MFVI (VCL) forgets tasks---any hyperparameter configuration that allows it to fully learn the most recent task makes it forget old tasks completely. The shown run offers the best average accuracy over all tasks for the last model. Our Radial BNN preserves information somewhat better even while training to a higher accuracy on the final task. Average accuracy is the dotted line.}
    \vspace{-2mm}
\label{fig:single_vcl}
\end{figure}

Our Radial BNN does not solve the continual learning single-headed problem, but it does show improved performance relative to the MFVI baseline. 
As we show in Figure \ref{fig:single_vcl}, the Radial BNN shows some remembering on old tasks (which includes identifying the task that the image comes from).
Moreover it is able to maintain good accuracy on the newest task.
Meanwhile, the hyperparameters that allow MFVI to optimize last-task average accuracy mean it learns a very uncertain model which has bad accuracy on the newest tasks.
This is because hyperparameters that would let it learn a high-accuracy model for the newest task would cause it to forget everything it saw earlier.

\section{Results on the UCI datasets}\label{a:uci}
We do not believe that the standard UCI Bayesian learning experiments which are heavily used in the field offer much insight in this case.
This is because all of the problems have low dimension (4-16) and because the experimental design allows only for a single hidden layer with 50 units.
This is required because many of the expensive techniques that researchers develop and evaluate on the UCI datasets only scale to very small models and inputs.

For sake of completeness we show some results of our methods on the UCI datasets.
As expected, our method does not outperform the more expensive techniques with complex covariances within the approximate posterior.
Moreover, as expected our method performs very similarly to MFVI.
In the small number of parameters involved in the UCI dataset experimental settings, the sampling problems for MFVI do not become severe.
In this low-dimensional regime, we do not expect any particular advantage of Radial BNNs over MFVI, which is what we find.

Note that in some cases the MFVI and Radial BNN results we show are somewhat worse than those reported in other papers.
We believe this is because of the fact that the resources devoted to hyperparameter search are not always the same in different papers.
We only searched over learning rates of 0.001 and 0.0001 using Adam and batch sizes of 16, 64, and 1000.
In the majority of datasets listed our results are competitive with what previous authors report for MFVI.
\begin{table*}[t]
\centering
\resizebox{\textwidth}{!}{%
\begin{tabular}{lccccccccc}
    
 \textbf{Dataset} &  \multicolumn{1}{c}{\textbf{MFVI}} & \multicolumn{1}{c}{\textbf{Radial}} & \multicolumn{1}{c}{\textbf{Dropout}} & \multicolumn{1}{c}{\textbf{VMG}}& \multicolumn{1}{c}{\textbf{FBNN}} & \multicolumn{1}{c}{\textbf{PBP\_MV}} & \multicolumn{1}{c}{\textbf{DVI}}\\\hline
 & \multicolumn{7}{c}{Avg. Test LL and Std. Errors}\\ \hline
Boston & -2.58$\pm$0.06 & -2.58$\pm$ 0.05 & -2.46$\pm$0.25 & -2.46$\pm$0.09 & -2.30$\pm$0.04 & 02.54$\pm$0.08 & -2.41$\pm$0.02 & \\
        Concrete &-5.08$\pm$0.01 &-5.08$\pm$0.01 & -3.04$\pm$0.09&-3.01$\pm$0.03& -3.10$\pm$0.01 & -3.04$\pm$0.03 & -3.06$\pm$0.01\\
        Energy & -1.05$\pm$0.01& -0.91$\pm$0.03 & -1.99$\pm$0.09 & -1.06$\pm$0.03 & \textbf{-0.68$\pm$0.02} & -1.01$\pm$0.01 & -1.01 $\pm$ 0.06\\
        Kin8nm  & 1.08$\pm$0.01 & \textbf{1.35$\pm$0.00} & 0.95$\pm$0.03 &1.10$\pm$0.01 & - & 1.28$\pm$0.01 & 1.13$\pm$0.00\\
        Naval  &-1.57$\pm$0.01& -1.58$\pm$0.01 & 3.80$\pm$0.05& 2.46$\pm$0.00 & \textbf{7.13$\pm$0.02}& 4.85$\pm$0.06 & 6.29$\pm$0.04\\
        Pow. Plant  & -7.54$\pm$0.00 & -7.54$\pm$0.00 & -2.80$\pm$0.05 &-2.82$\pm$0.01 & - & -2.78$\pm$0.01 & -2.80$\pm$0.00\\
       Protein &-3.67$\pm$0.00&-3.66$\pm$0.00&-2.89$\pm$0.01 &-2.84$\pm$0.00 & -2.89$\pm$0.00 & -2.77$\pm$0.01 & -2.85$\pm$0.01\\
        Wine &-3.15$\pm$0.01 & -3.15$\pm$0.01 & -0.93$\pm$0.06 & -0.95$\pm$0.01 & -1.04$\pm$0.01 & -0.97$\pm$0.01 & \textbf{-0.90$\pm$0.01}\\
        Yacht &-4.20$\pm$0.05 &-4.20$\pm$0.05& -1.55$\pm$0.12&-1.30$\pm$0.02 & -1.03$\pm$0.03 & -1.64$\pm$0.02 & \textbf{-0.47$\pm$0.03}\\ \hline
        & \multicolumn{7}{c}{Avg. Test RMSE and Std. Errors} \\\hline
Boston & 3.42$\pm$0.23 & 3.36$\pm$0.23 & 2.97$\pm$0.85 &2.70$\pm$0.13 & 2.38$\pm$0.10 & 3.11$\pm$0.15 & -\\
        Concrete &5.71$\pm$0.15& 5.62$\pm$0.14 & 5.23$\pm$ 0.53 &4.89$\pm$0.12 & 4.94$\pm$0.18& 5.08$\pm$0.14& -\\
        Energy & 0.81$\pm$0.08& 0.66$\pm$0.03 & 1.66$\pm$0.19 & 0.54$\pm$0.02 & 0.41$\pm$0.20 &0.45$\pm$0.01 & -\\
        Kin8nm  & 0.37$\pm$0.00 & 0.16$\pm$0.00& 0.10$\pm$0.00&0.08$\pm$0.00 & - & \textbf{0.07$\pm$0.00} &-\\
        Naval & 0.01$\pm$0.00 & 0.01$\pm$0.00 & 0.01$\pm$0.00 &0.00$\pm$0.00 & 0.00$\pm$0.00 & 0.00$\pm$0.00&-\\
        Pow. Plant  &4.02$\pm$0.04 & 4.04$\pm$0.04 & 4.02$\pm$0.18 & 4.04$\pm$0.04 & - & 3.91$\pm$0.14&-\\
        Protein & 4.40$\pm$0.02 & 4.34$\pm$0.03 & 4.36$\pm$0.04 & 4.13$\pm$0.02& 4.33$\pm$0.03& 3.94$\pm$0.02&-\\
        Wine & 0.65$\pm$0.01 & 0.64$\pm$0.01 & 0.62$\pm$0.04 & 0.63$\pm$0.01 & 0.67$\pm$0.01 & 0.64$\pm$0.01&-\\
        Yacht &1.75$\pm$0.42& 1.86$\pm$0.37&1.11$\pm$0.38&0.71$\pm$0.05& \textbf{0.61$\pm$0.07} & 0.81$\pm$0.06&-\\
    \end{tabular}%
   }
    \caption{Avg.\ test RMSE, predictive log-likelihood and s.e.\ for UCI regression datasets. Bold where one model is better than the next best $\pm$ their standard error. Results are from multiple papers and \textit{hyperparameter search is not consistent}. MFVI and Radial are our implementations of standard MFVI and our proposed model respectively. Dropout is \citet{gal_dropout_2015}. Variational Matrix Gaussian (VMG) is \citet{louizos_structured_2016}. Functional Bayesian Neural Networks (FBNN) is \citet{sun_functional_2019}. Probabilistic Backpropagation Matrix Variate Gaussian (PBP\_MV) is \citet{sun_learning_2017}. Deterministic VI (DVI) is \citet{wu_deterministic_2019}.}
    \label{tab:reg_res}
    \vspace{4mm}
\end{table*}

\section{Further Gradient Experiment}\label{a:further_grad_exp}
\begin{figure}
\vspace{-1mm}
\includegraphics{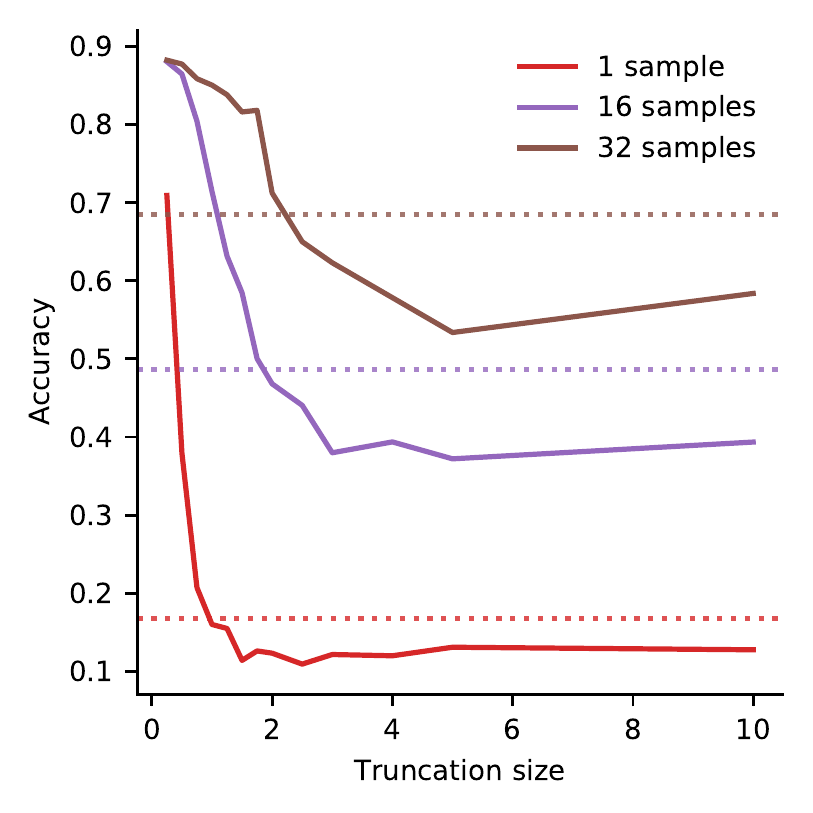}
\vspace{-8mm}
\caption{Dotted lines show untruncated Gaussian performance. Highly truncated Gaussians \textit{improve} MFVI. This effect is most significant when small numbers of samples from the posterior are used to estimate the gradient. We conclude that despite bias, the low variance offered by truncation improves gradient estimates. Results averaged over 10 initial seeds for each truncation size.} \label{fig:truncation}
\end{figure}

A further analysis demonstrates that we can improve the performance of MFVI models by using a low-variance but highly biased estimator of the NLL loss.
We do this by estimating the NLL with a truncated version of the Gaussian sampling distribution, without changing how we analytically calculate the KL terms of the loss.
We use rejection sampling, selecting only samples from a Gaussian distribution which fall under a threshold.

Our new estimate of the loss is biased (because we are not sampling from the distribution used to compute the KL divergence) but has lower variance (because only samples near the mean are used).

In Figure \ref{fig:truncation} we show that the truncated models to outperform `correct' MFVI with standard deviations initialized slightly too high (we used $\sigma=0.12$).
This is despite the fact that we are using a biased estimator.
This supports the hypothesis that MFVI training is hamstrung by high gradient variance.

Moreover, the smaller the number of samples, the higher the variance of the estimator will be, and the bigger a problem we might expect variance to be for training.
Indeed, we show that the effect of truncation is smaller for larger numbers of samples.
This suggests that estimating the gradient of the loss function for MFVI is hampered by sampling far from the mean, and that this effect is linked to the variance of estimates of the gradient.

\end{document}